\documentclass[10pt,twocolumn,letterpaper]{article}

\usepackage[pagenumbers]{cvpr} % To force page numbers, e.g. for an arXiv version

\usepackage{graphicx}
\usepackage{amsmath}
\usepackage{amssymb}
\usepackage{booktabs}
\usepackage{mathtools}
\usepackage{tabularx}
\usepackage{multirow}
\usepackage[pagebackref,breaklinks,colorlinks]{hyperref}

\usepackage[capitalize]{cleveref}
\crefname{section}{Sec.}{Secs.}
\Crefname{section}{Section}{Sections}
\Crefname{table}{Table}{Tables}
\crefname{table}{Tab.}{Tabs.}

\def\cvprPaperID{6018} % *** Enter the CVPR Paper ID here
\def\confName{CVPR}
\def\confYear{2023}

\begin{document}

%%%%%%%%% TITLE - PLEASE UPDATE
\title{Visibility-Aware Pixelwise View Selection for Multi-View Stereo Matching}

\author{Zhentao Huang\\
University of Guelph\\
{\tt\small zhentao@uoguelph.ca}
% For a paper whose authors are all at the same institution,
% omit the following lines up until the closing ``}''.
% Additional authors and addresses can be added with ``\and'',
% just like the second author.
% To save space, use either the email address or home page, not both
\and
Yukun Shi\\
University of Guelph\\
{\tt\small yshi21@uoguelph.ca}
\and
Minglun Gong\\
University of Guelph\\
{\tt\small minglun@uoguelph.ca}
}
\maketitle

%%%%%%%%% ABSTRACT
\begin{abstract}
   The performance of PatchMatch-based multi-view stereo algorithms depends heavily on the source views selected for computing matching costs. Instead of modeling the visibility of different views, most existing approaches handle occlusions in an ad-hoc manner. To address this issue, we propose a novel visibility-guided pixelwise view selection scheme in this paper. It progressively refines the set of source views to be used for each pixel in the reference view based on visibility information provided by already validated solutions. In addition, the Artificial Multi-Bee Colony (AMBC) algorithm is employed to search for optimal solutions for different pixels in parallel. Inter-colony communication is performed both within the same image and among different images. Fitness rewards are added to validated and propagated solutions, effectively enforcing the smoothness of neighboring pixels and allowing better handling of textureless areas. Experimental results on the DTU dataset show our method achieves state-of-the-art performance among non-learning-based methods and retrieves more details in occluded and low-textured regions.      
\end{abstract}

%%%%%%%%% BODY TEXT

\section{Introduction}

Multi-view stereo (MVS),  which estimates dense 3D point clouds from a set of calibrated input images, is an important research topic and supports many downstream applications, such as autonomous driving, 3D reconstruction, and virtual reality. Even though much progress has been made in recent years \cite{furukawa2009accurate, schonberger2016pixelwise, tola2009daisy, campbell2008using}, reconstructing accurate and complete 3D point cloud models remains challenging due to obstacles caused by low texture, reflections, occlusions, and repetitive patterns.

Inspired by the success of MVSNet \cite{yao2018mvsnet}, numerous learning-based methods \cite{luo2019p, yao2019recurrent,gu2020cascade,wang2021patchmatchnet,peng2022rethinking} had been proposed in recent years and shown outstanding performances. They had been ranked on the top of various MVS datasets \cite{aanaes2016large, yao2020blendedmvs, knapitsch2017tanks}. However, it is questionable how well these learning-based methods can adapt to scenes underrepresented in the training dataset and how much time these models take for training and fine-tuning on a new dataset.

Recently, PatchMatch-based methods \cite{galliani2015massively, xu2019multi, xu2020planar, schonberger2016pixelwise} also show excellent capability in depth map estimation. Following \cite{bleyer2011patchmatch}, these methods generally have a four-step pipeline: random initialization, propagation, view selection, and refinement. View selection is an important factor here because correct matches can only be found from nearby unoccluded views, and occlusions are common under the MVS setting. Yet, existing approaches often resort to ad-hoc view selection methods (\eg top-n views with the lowest matching cost \cite{galliani2015massively}) without considering visibility constraints. Therefore, a motivating question is whether we can make the view selection process visibility-aware and how much benefit we can gain from such enhancement.

\begin{figure*}
  \centering
    \includegraphics[width=\linewidth]{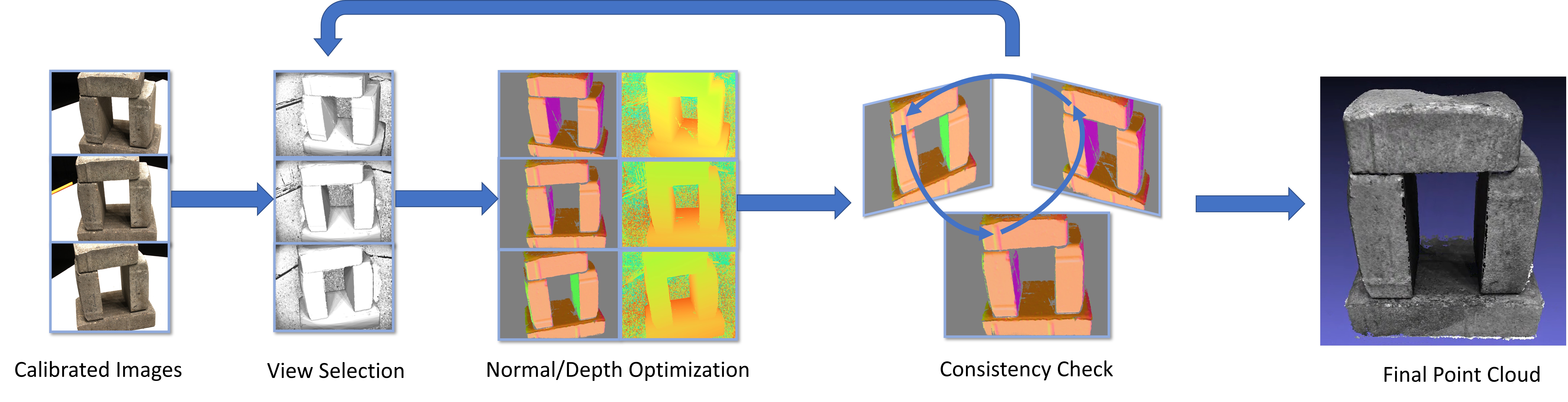}
    \caption{Overview of the proposed algorithm, which iteratively performs view selection, normal and depth optimization, and geometry consistency check. Once the process converges, 3D point clouds estimated from different views are fused together to produce the final model. 
    %manner. The results are fine-tuned with the adoption of geometric constraints over cycles. (a) Initial surface normal, depth, and visibility estimation by AMBC (Section \ref{sec:ambc}). (b) Surface hypothesis verified by Geometric Consistency Check (Section \ref{sec:gcc}). (c) Surface normal, depth, and visibility estimation by AMBC with geometric constraints. (d) Depth map and normal map fusion (Section \ref{sec:fusion}).
    % \mlc{update the figure to show view-selection.  the loop should be from input calibrated (not `rectified') images -> view selection -> normal/depth optimization -> consistency check -> back to view selection.  Also show visibility maps as view selection maps.}
    }
    \label{fig:overview}

  \hfill
\end{figure*}

To this end, we develop a pixelwise view selection approach, which progressively updates the source views used for each pixel. The selected views will be used for both matching cost calculation and depth/normal consistency check, which leads to a set of validated solutions. These validated solutions are used to guide future view selections through visibility checks; see Figure \ref{fig:overview}.

%both for multi-view stereo matching. Our key idea is to perform view selection from the geometric level. There are reasons for unreliable views: the view is occluded, the view is too far from the reference view, the angle between the surface normal and the viewing ray is too large, etc. By applying constraints to address these situations, our method aggregates all credible views. However, it still encounters failure in large regions with low texture. Our observation is that true and false hypotheses both have relatively low matching costs in the low-textured areas. Therefore, we first apply strong and weak geometry consistency checks to verify the existing solutions. Then apply smoothness constraint by adding a reward to those verified solutions when propagating. Ideally, smoothness constraint will break local optimum in low-textured areas and not affect high-textured areas, where true and false hypotheses have a relatively large difference.

Even with a proper set of source views selected, searching the optimal depth and normal for each pixel is still a challenging problem. We address this issue using three strategies: 1) A swarm-based optimization framework (the Artificial Multi-Bee Colony algorithm \cite{wang2016artificial}) is utilized to avoid being trapped into local optimum; 2) Both intra-image and inter-image solution propagation are employed to speed up the convergence; 3) A smoothness term is added into intra-image propagation to handle low texture areas better. 

%Artificial Multi-Bee Colony Algorithm. The original Artificial Bee Colony algorithm is proposed by \cite{karaboga2005idea}, then extended by \cite{wang2016artificial} for the K-Nearest-Neighboring Fields search. It shows great global searching power with a strong capability to avoid local optimum.

% Our main contributions are summarized as follows:
% \begin{itemize}
% 	\item  We propose a novel progressive pixelwise view selection method, which is visibility-aware. It supports both matching cost calculation and consistency check.
	
% 	\item Inspired by \cite{wang2016artificial}, we applied the Artificial Multi-Bee Colony (AMBC) algorithm to multi-view stereo. The between-colony onlooker bee is used for both intra-image and inter-image solution propagation.
	
% 	\item To better handle low-textured areas, fitness rewards are introduced when propagating intra-image solutions validated through consistency check. This allows smoothness constraint to be enforced efficiently and effectively. 

% \end{itemize}

\section{Related Work}

Related works in the MVS field can be mainly divided into two aspects, 
learning-based and non-learning-based (traditional MVS).

\paragraph{Learning-based Shape Reconstruction.}
The huge success of deep neural networks in image classification has sparked interest in introducing learning to other domains. For the 3D shape reconstruction problem, a natural path for applying the standard convolutional neural network (CNN) architecture is to estimate a 2.5D depth map for each input image or to operate on 3D voxels. Depth map based approaches are proposed to infer per-pixel depth either from single image\cite{liu2015learning} or to use visual cues extracted from stereo pairs \cite{godard2017unsupervised, yin2018geonet}. Additional processing is then needed to consolidate multiple depth maps into a single 3D point cloud\cite{wu2017structure, mao2022robust}. Voxel-based methods\cite{choy20163d, tatarchenko2017octree} utilize 3D convolution operators to encode and decode geometric features in discretized 3D space directly but are limited to the relatively low voxel resolution. To address these limitations, 3D point cloud based\cite{thomas2019kpconv, yang2019pointflow} and implicit surface based\cite{chen2019learning, mescheder2019occupancy, sitzmann2020metasdf} approaches are also proposed. Latter research finds that implicit surfaces can be approximated by fully connected networks\cite{niemeyer2020differentiable}, which leads to reconstruction using neural implicit representation\cite{wei2021nerfingmvs}.

%Since surface models are widely used for rendering and animating 3D objects, additional steps are often needed to convert 3D point clouds or voxels into meshes. There are growing interests in generating polygon meshes for input scenes directly. Both implicit\cite{chen2020bsp, deng2020cvxnet} and explicit\cite{groueix2018papier} approaches are developed. Typically, these methods train a binary classifier to approximate the border of the surface during training, which enables to the formulation of 3D polygon meshes during inferencing.

% In recent year, deep neural networks (DNNs) shows outstanding performance in various computer vision tasks, such as ... %related applications
% Learning-based Multi-view Stereo approaches can also be divided into volumetric-based method and depth map-based method according to scene representation. 

%related learning-based papers
% SurfaceNet presented a neural network that reconstructed the 2D surface \cite{ji2017surfacenet}. It is an end-to-end framework that directly infers the 3D model based on a set of images and their 
% corresponding camera parameters. 

\paragraph{Non-learning-based Multi-view Stereo.}
Non-learning-based MVS computes the matching cost of rectified image patches by different methods, 
such as the Sum of Absolute Differences (SAD), Sum of Squared Distances (SSD), 
or Normalized Cross-Correlation (NCC). 
According to\cite{seitz2006comparison}, generally, there are four types of methods used to represent scene: 
volumetric based\cite{kutulakos2000theory, seitz1999photorealistic},
point cloud based\cite{furukawa2009accurate, lhuillier2005quasi},
mesh based\cite{tang2019skeleton, tang2021skeletonnet},
and depth map based\cite{galliani2015massively, xu2019multi}. 
Recently, due to parallel ability and high performance, 
PatchMatch\cite{galliani2015massively, xu2019multi} is widely used in this field.
The core idea of Patchmatch\cite{barnes2009patchmatch} is to minimize the matching cost of every pixel and the plane in disparity space and then effectively estimate depth maps for every image. \cite{schonberger2016pixelwise} estimates depth maps and pixelwise view selection jointly. 

% \textbf{PatchMatch} 

\paragraph{Depth Map Fusion.}
There are two main types of Multi-view Stereo approaches categorized according to the scene representation: volumetric-based methods and depth map-based methods \cite{wang2021multi}. The depth map representation provides pixel-wise depth information for every viewpoint. The 3D reconstruction represented by the point cloud can be recovered by applying 3D fusion techniques to all the depth maps. Unlike depth map-based approaches, which construct the scene indirectly, volumetric representation depicts the scene based on volume occupancy in 3D space. \cite{sinha2007multi} proposed a graph cuts-based reconstruction that generates multi-resolution volumetric meshes.

\paragraph{Artificial Multi-Bee-Colony Algorithm.}
Artificial Bee Colony (ABC) algorithm\cite{karaboga2005idea} and its variants\cite{ZHU20103166} are used to handle constrained and unconstrained optimization problems\cite{karaboga2007artificial}. Compared with other population-based algorithms, the ABC algorithm can achieve equal or better performance with fewer parameters. Thus, the ABC algorithm has been used in many applications such as feature selection\cite{hancer2015binary} and data clustering\cite{tan2014improved}.

Artificial Multi-Bee-Colony (AMBC) algorithm\cite{wang2016artificial} searches k-nearest matches via dedicated bee colonies. The communication in colonies can propagate proper matches to escape local optima. AMBC is also able for parallel processing as it makes no assumption about the neighbor or direction.

Our work leverages the idea proposed by AMBC to build the MVS framework. We not only apply the between-colony communication idea for propagating solutions between different pixels of the same image but also for propagation among different images. In addition, rewards are added to validated intra-image solution propagation, which allows simple yet effective enforcement of smoothness constraint.
%As far as we know, we are the first to apply the AMBC algorithm to optimize MVS problem. In addition, the proposed method belongs to the depth map-based approach, and the related PatchMatch Stereo approaches will be mainly discussed in the rest of the section. 
\section{Overview}

Given a set of input 2D images $I = \{I_i | i = 1\cdots N\}$ with known camera parameters $C = \{C_i | i = 1\cdots N\}$, the goal of MVS is to estimate pixel-wise depth maps $d = \{d_i | i = 1\cdots N\}$ for every view and fuse them into a 3D point cloud. Specifically, when processing a reference image $I_{\mathit{ref}}$, MVS algorithms normally estimate a local fitting plane $P$ for each pixel $x$ in $I_{\mathit{ref}}$'s local coordinates, using some of the remaining views as source images $I_{src} \in \{I\} - I_{\mathit{ref}}$. The plane $P$ depicts both the depth and normal information of the local geometry, which are denoted as $d_{\mathit{ref}}(x)$ and $\vec{n}_{\mathit{ref}}(x)$, respectively.

Figure \ref{fig:overview} illustrates an overview of our method. We construct a three-phased process that evolves over cycles. An initial set of source views is selected for each pixel in each reference image based on only camera parameters $C$. These source views are used to compute matching costs, based on which an AMBC algorithm is used to search for the optimal solution (depth and normal) for each pixel. A geometry consistency check is then performed by projecting the optimal solution found for a given pixel to all of its source views. The solution is considered validated if both depth and normal are consistent with the corresponding pixels in these source views. Validated solutions are further used for: 1) guiding the future source view selection process as some source views may be determined as occluded; 2) communicating both among pixels in the same image and among different images; and 3) fusing into the final point cloud.

\section{Pixelwise View Selection}\label{viewselection}

The selection of source views for matching cost calculation strongly impacts the quality of reconstruction results. Previous work has proposed to use triangulation angle, incident angle, and image resolution-based geometric priors to perform pixelwise view selection \cite{schonberger2016pixelwise}. While we acknowledge the importance of resolution-based geometric prior in handling large-scale scenes, its benefit for reconstructing a small set of objects is limited since all input images have similar resolutions. Hence, we removed this term in our implementation for simplicity. Instead, we add a visibility-based term, which handles occlusions based on geometric information instead of heuristics.

\begin{figure*}
        \includegraphics[width=\linewidth]{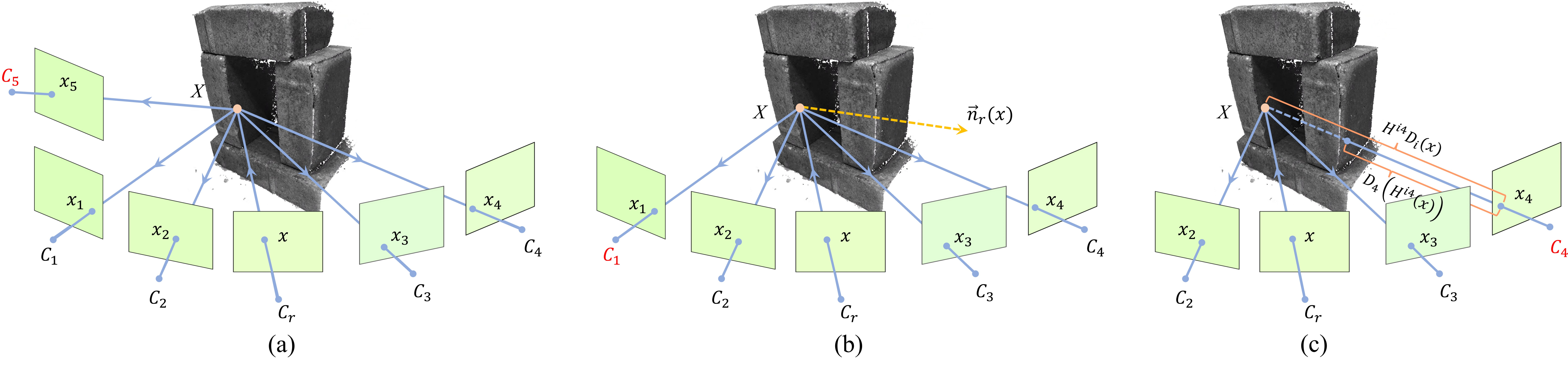}
        \caption{Illustration for the progressive view selection process: (a) initially, only source views with poor triangulation angle ($C_5$) will be filtered out from the set;
        (b) once the normal for $x$ is estimated, views with poor incident angles ($C_1$) will be removed from the source view set; and (c) validated solutions are used to further remove occluded views ($C_4$) from the source view set.}\label{fig:viewselect}
\end{figure*}

The actual terms used for view selection are based on available information. At the beginning of the process, we have yet to gain prior knowledge of scene geometry. Therefore, when processing image $I_i$ as the reference view, only the triangulation angle term is used for view selection. All nearby views whose triangulation angle with $I_{i}$ is between $[10^{\circ},30^{\circ}]$ are selected into the source view set $\{I_{src}\}$; see Figure \ref{fig:viewselect}(a). It is worth noting that the same set of views is used for all pixels in $I_{i}$.

Once the depth $d_i(x)$ and normal $\vec{n}_i(x)$ for each pixel $x$ in $I_{i}$ are estimated, the incident angle term will be used. If a given view $I_j$ has a poor incident angle, \ie the angle between $\vec{n}_i(x)$ and the viewing vector of $I_j$ is greater than $80^{\circ}$, $I_j$ will be removed from the source view set $\{I_{src}\}$; see Figure \ref{fig:viewselect}(b). As a result, the set $\{I_{src}\}$ will be adaptively determined for different pixels in $I_{i}$.

Finally, once validated depth and normal are found for different views (details on solution validation will be discussed in Sec.~\ref{sec:gcc}), the visibility term will be introduced. That is, for a given pixel $x$ in reference view $I_{i}$, we will first backproject $x$ to a 3D scene point $X$ using the estimated depth $d_i(x)$. The 3D point $X$ is then projected to each view $I_j$ in set $\{I_{src}\}$. Without losing generality, here we assume the projection of $X$ on image $I_j$ is pixel $y$. We consider the $X$ is occluded in $I_j$ if and only if a validated solution is found at pixel $y$ and the depth $d_j(y)$ is smaller than the distance to 3D point $X$. $I_j$ will be removed from source view set $\{I_{src}\}$ if $X$ is occluded in $I_j$; see Figure \ref{fig:viewselect}(c).

%The first constraint addresses the occlusion situation, while the occluded view could not contribute or even have a negative effect on cost aggregation. The second constraint is applied because when the angle between the viewing ray and the surface normal is too large, small variations in pixels will result in a large difference in depth. Instead of heuristically defining how many neighboring views could contribute to the cost, we use every view that could correctly capture the point at the geometric level. With pixel-wise view selection, we are also able to generate a visibility map to illustrate how many neighboring views are contributing to the cost computation at each pixel.

\begin{figure*}
        \begin{subfigure}[]{0.25\textwidth}
                \includegraphics[width=\linewidth]{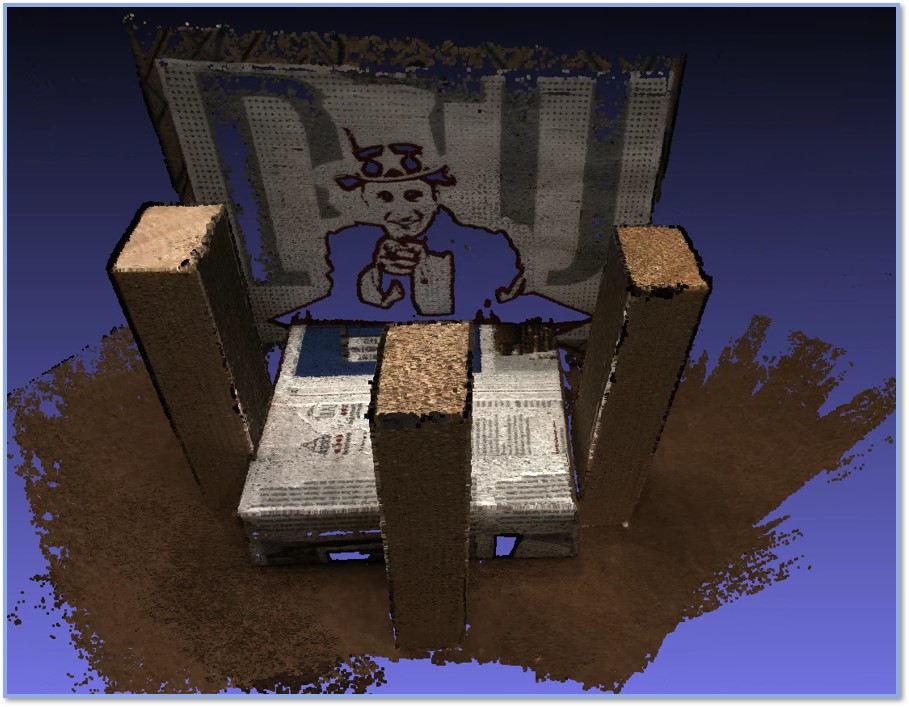}
                \caption{COLMAP \cite{schonberger2016pixelwise}}
                \label{fig:gull}
        \end{subfigure}%
        \begin{subfigure}[]{0.25\textwidth}
                \includegraphics[width=\linewidth]{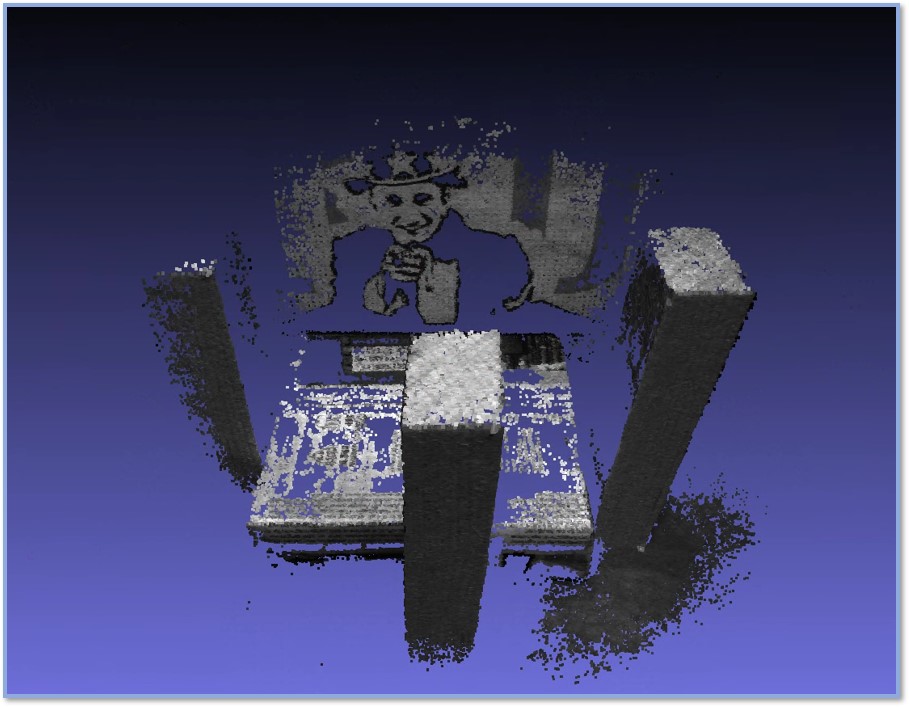}
                \caption{Gipuma \cite{galliani2015massively}}
                \label{fig:gull2}
        \end{subfigure}%
        \begin{subfigure}[]{0.25\textwidth}
                \includegraphics[width=\linewidth]{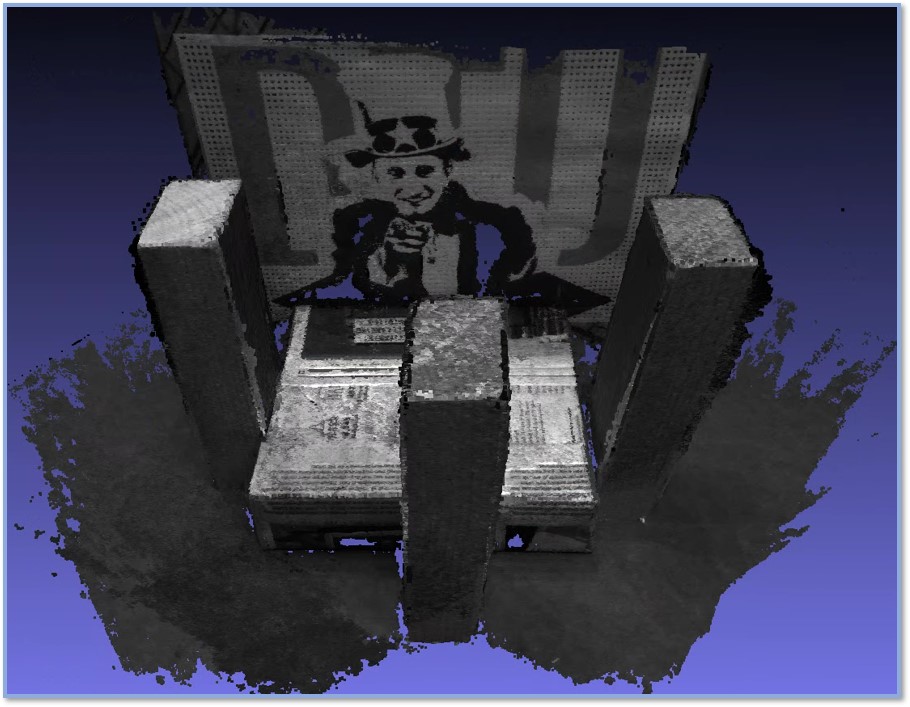}
                \caption{Ours}
                \label{fig:tiger}
        \end{subfigure}%
        \begin{subfigure}[]{0.25\textwidth}
                \includegraphics[width=\linewidth]{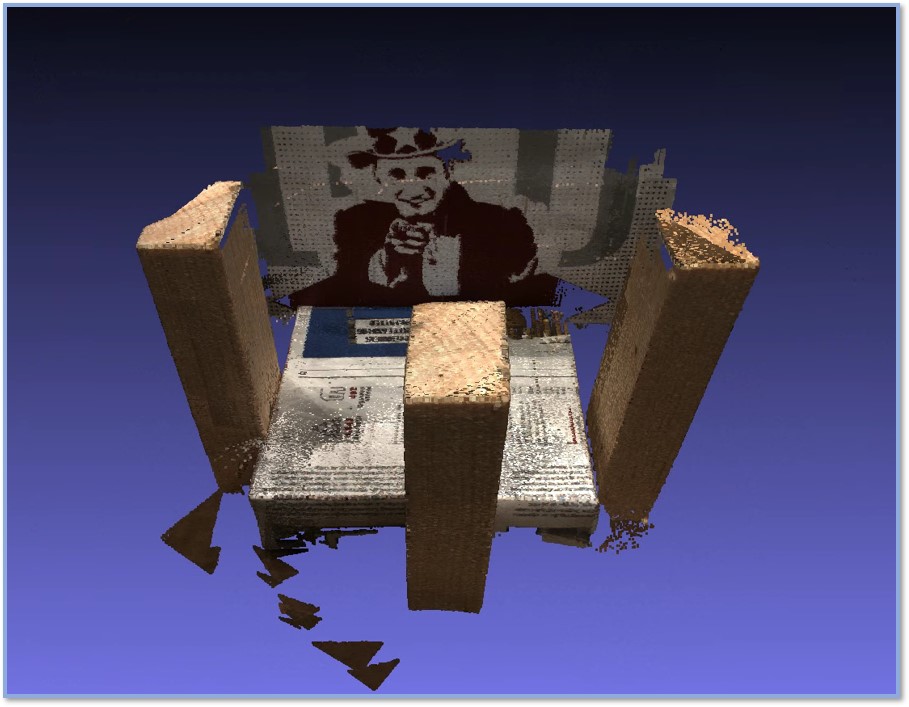}
                \caption{Ground Truth}
                \label{fig:mouse}
        \end{subfigure}
        \caption{Comparison of the reconstructed point cloud on DTU dataset. Our approach outperform in less-textured regions, thanks to the added smoothness reward term.}\label{fig:animals}
\end{figure*}

\begin{figure*}
    \centering
    \includegraphics[width=\textwidth]{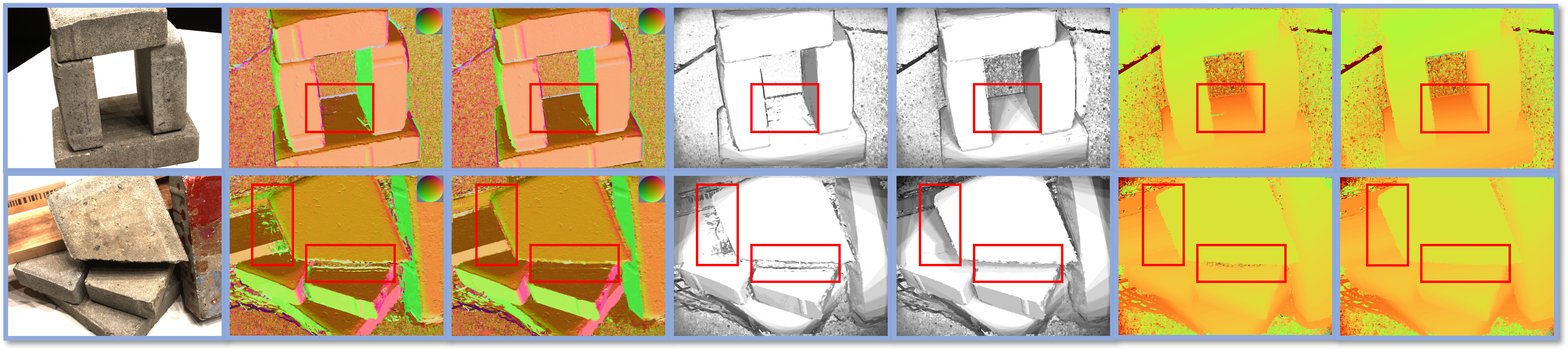}
    \caption{An illustration on the effect of Pixelwise View Selection (PVS). From left to right: input image, normal map without/with PVS, source view set map without/with PVS (brightness indicates the number of views in the source view set), depth map without/with PVS. Challenging areas are marked in red boxes.}
    \label{fig:pvs}
\end{figure*}

\section{Artificial Multi-Bee Colony Algorithm}\label{sec:ambc}
The original Artificial Bee Colony algorithm is a bio-inspired optimization approach proposed by Karaboga \cite{karaboga2005idea}. Wang \etal \cite{wang2016artificial} later employ multiple bee colonies to optimize solutions for different pixels and use inter-colony scout bees to facilitate solution propagation.

In MVS context, each solution has four parameters: three for the normal vector and one for the depth value. By representing each solution as a food source, the optimal results are searched by sending out three kinds of bees: employed bees, onlooker bees, and scout bees. The parameterization is conducted in the Euclidean scene space as Gipuma \cite{galliani2015massively} does. Compared to the disparity space, parameterization in scene space avoids the epipolar rectification, and it could generate dense normals in the scene, which could be used for further point cloud fusion \cite{kazhdan2013screened}. 

In the Euclidean scene space, the plane equation for 3D object points $\textbf{X} =[X,Y,Z]^\top$ would be $\vec{n}^\top \textbf{X}=-D$, where $\vec{n}$ is the normal vector and $D$ is the distance to the origin. By placing the reference camera at the origin, the depth $d$ at the pixel $x$ could be inferred with the plane parameters and the camera intrinsic parameters:

\begin{equation}
d = \frac{D f_x}{[x-u, f_x/f_y(y-v), f_x]\cdot \vec{n}}\label{eq:depth}
\end{equation}
%   K = 
%  \begin{bmatrix}
% f_x & 0 & c_x\\
% 0 & f_y & c_y\\
% 0 & 0 & 1
% \end{bmatrix}
In this equation, $c_x,c_y$ represent the optical center, and $f_x, f_y$ represent the focal length of the camera in pixels which are parameters from camera intrinsic matrix $K$. Under this context, each solution contains four parameters: three for $n$ and one for $d$. Then the pixel $x$ in the reference image $K[I|0]$ is related to the corresponding point $x'$ in the source image $K'[R|t]$ based on the plane-induced homography \cite{hartley2003multiple}:

\begin{equation}
    H_\pi = K'(R-\frac{1}{D}t\vec{n}^\top)K^{-1}
\end{equation}

 \begin{figure}[t]

    \includegraphics[width=\linewidth]{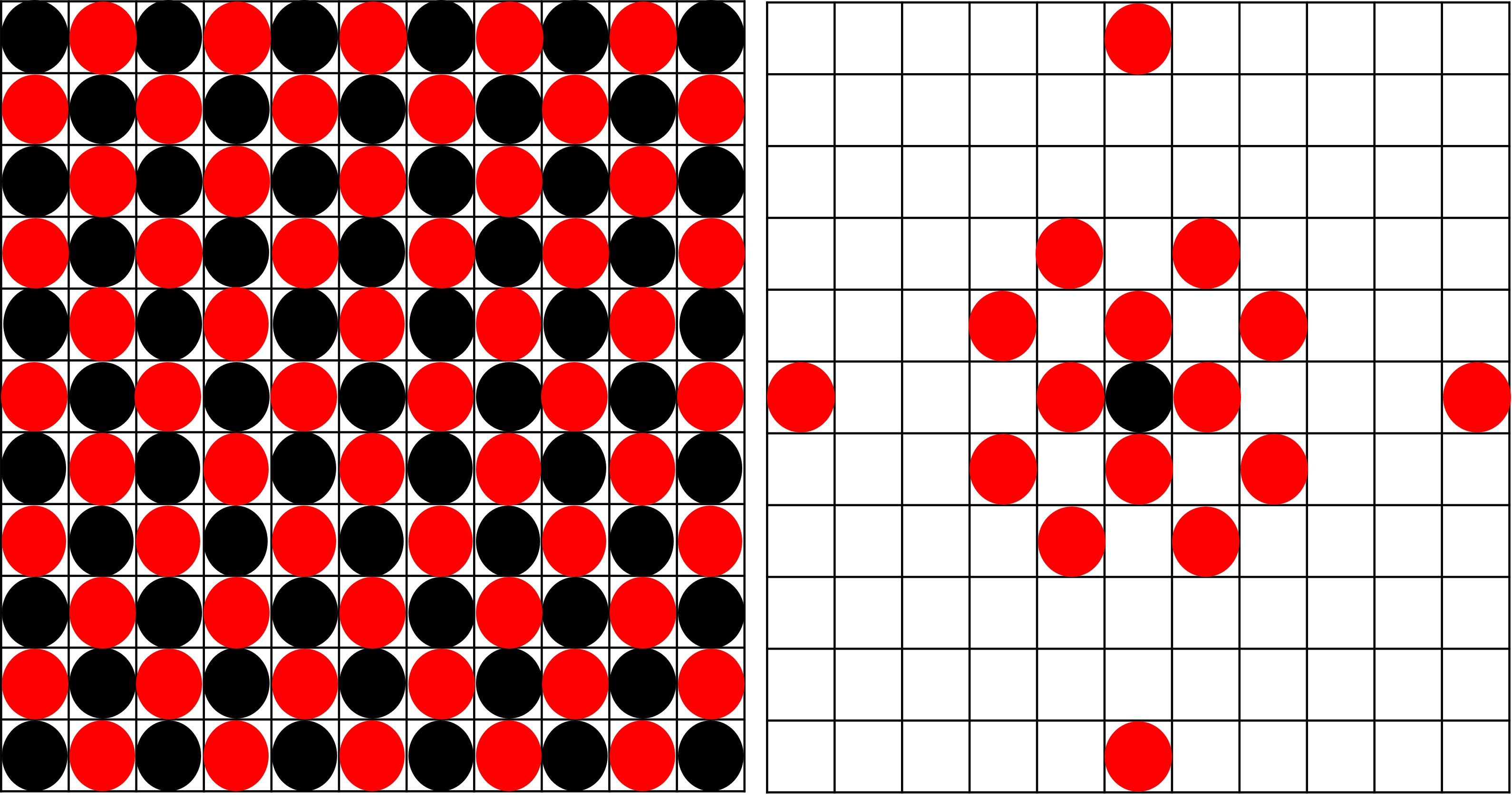}
    \caption{Left: red-black checkerboard pattern. Each block represents a pixel in the image. The pixels in the same color are processed parallelly. Right: the sample space (red) of a center pixel (black).}
    \label{fig:checkerboard}

  \hfill
 
\end{figure}

%overview of AMBC algorithm

\subsection{Random Initialization}\label{sec:init}

We first randomly generate the same number of hypotheses as preset $\mathit{FoodNumber}$ for each pixel. Each hypothesis contains three parameters for the normal vector $\vec{n}$, and one parameter for plane depth $D$. We follow \cite{marsaglia1972choosing} to uniformly sample the normal vector over the visible hemisphere. Note that since plane depth $D$ is used here, the solutions of pixels on the same plane will have the same set of parameters. Scene depth $d$ could be obtained via equation \ref{eq:depth} when needed.

% \begin{equation}
%     \textbf{n} = [1-2S, 2q_1\sqrt{1-S}, 2q_2\sqrt{1-S}]^\top
%     \label{eq:random_normal}
% \end{equation}

% If the normal vector has the same direction as the principal ray, it is then inverted. Following \cite{galliani2015massively}, the last parameter is uniformly sampled in the disparity space rather than scene space and then transformed to $d$ by Equation \ref{eq:depth}. It will produce a more densely set of depth from the near field and a sparser set in the far field. %It could save unnecessary calculation consumption in the far field where small variations do not make an observable difference.

\subsection{Matching Cost Evaluation}

The similarity between two patches related via plane-induced homography defines whether the hypothesis depicts the scene correctly. In the two-view stereo, the matching cost is straightforward. When extending to multi-view stereo, we adopt the following equation to aggregate the matching cost of hypothesis $\langle d_i(x), \vec{n}_i(x) \rangle$ in pixel $x$ of the reference image $i$:

\begin{equation}
    C(d_i(x), \vec{n}_i(x))= 
\begin{dcases}
    \frac{\sum\limits_{j\in \{S_{src}\}} m(i,j)}{|\{S_{src}\}|-1},& \text{if } |\{S_{src}\}| > 1\\
    +\infty,              & \text{otherwise}
\end{dcases}
\label{eq:cost}
\end{equation}
where $m(i,j)$ represents the matching cost between two patches from reference view $i$ and source view $j$, and $\{S_{src}\}$ is the set of suitable source views for pixel $x$. In this paper, we adopt bilaterally weighted Normalized Cross-Correlation \cite{schonberger2016pixelwise} as our matching cost function.

%The most basic type of valid set $S_{valid}$ is selected based on two criteria: First, the center rays of reference view and source view should have a relatively small angle. This is used to reduce the calculation since most of the MVS datasets have tens or hundreds of images from different viewpoints. Second, the projected patch should be within the image boundary. These are two basic requirements for selecting valid source images for matching cost calculation. However, after the first cycle of depth-normal map estimation and the geometric consistency check, with more geometric constraints applied, only the source views with proper visibility of the current pixel will be selected as the valid set. Detailed information will be illustrated in Section \ref{viewselection}.

In Equation \ref{eq:cost}, the aggregation cost is divided by $|\{S_{src}\}|-1$ rather than $|\{S_{src}\}|$ because we expect the results to be an unbiased sample estimate that prefers a larger $\{S_{src}\}$ set. Which is more likely to capture the true normal and depth information than a small set where only several views produce the best results and the rest are invalid (\eg out of image boundaries).

In addition, we adopt a fitness value to each plane hypothesis:
\begin{equation}
    F(d_i(x), \vec{n}_i(x)) = \frac{1}{1+C(d_i(x), \vec{n}_i(x))}\label{eq:fitness}
\end{equation}
The fitness for a hypothesis is higher if the aggregation matching cost is lower. The trial count $T(d_i(x), \vec{n}_i(x))$ is added and set to zero for all food sources in the initialization. It is designed to track whether each food source is updated through iterations.

%  \begin{figure}[t]

%     \includegraphics[width=\linewidth]{cvpr2023-author_kit-v1_1-1/latex/imgs/strong.jpg}
%     \caption{Illustration of Strong Consistency Check. The solution in reference view $C_r$ is consistent with its neighboring views. The distance between 3D point $X$ and image plane $D_{p,1..4}$ equals to the depth value in pixels $x_{1..4}$. And the normal in pixel $x_r$ is also near to normal at $x_{1..4}$.}
%     \label{fig:strong}

%   \hfill
 
% \end{figure}
 
\subsection{Employed Bees}
The task for employed bees is to perform a search within the local colonies by randomly perturbing each food source. For each food source, the perturbed food source $y'_x$ is generated by:

\begin{equation}
    y'_x=y_x+R(-1,1)(y_n-y_x)
\end{equation}

where $R(-1,1)$ returns a value uniformly distributed between -1 and 1, and $y_n$ is another food source randomly selected within the colony. The fitness of the perturbed hypothesis is then evaluated based on the combination of Equation \ref{eq:cost} and \ref{eq:fitness}. If the fitness of perturbed food source $F(y'_x)$ is greater than $F(y_x)$, then $y_x$ is replaced by $y'_x$. Otherwise, set the trial count $T(x) = T(x) + 1$.

\subsection{Onlooker Bees}
 The task for onlooker bees is to perform searching in neighboring colonies. Following \cite{galliani2015massively}, we adopt a red-black checkerboard pattern for sampling. It divides the image into red and black groups. Pixels in the same color group can be processed in parallel without interfering with others. Figure \ref{fig:checkerboard} shows a center pixel's pattern and sample space.

For every food source $y_x$ with a black label, randomly select a colony with a red label following the pattern in Figure \ref{fig:checkerboard}, and vice versa. Then select the food source $y_n$ with the highest fitness value, and evaluate the fitness of the food source $F(y_n)$ in the current colony. If $F(y_n)$ is greater than $F(y_x)$, then $y_x$ is replaced by $y_n$. Otherwise, set the trial count $T(y_x) = T(y_x) + 1$.

\subsection{Scout Bees}

After several iterations after the initialization, the colony's food sources will converge to a relatively small range. Since employed bees and onlooker bees only perturb or copy the existing solutions, the task for scout bees is to perform global searching and avoid potential local optimum. In this paper, we only want to perform the global searching for $\mathit{FoodNumber} - 1$ food sources. The reason is that the food source with the highest fitness value should always be kept in the colony. Therefore, for the rest of the food source, if their trial count $T(y_x)$ exceeds a preset threshold (it is empirically set to 10), then it is replaced by a  randomly generated food source. The idea behind this is that when a food source has not been updated for certain iterations and is not the best in the colony, it may have been stuck in the local optimum. The new food source is evaluated via Equation \ref{eq:fitness}, and the new trial count is set to 0. 

\section{Geometric Consistency Check}
\label{sec:gcc}
Due to noise and/or deviation from Lambertian property, mismatches sometimes have lower matching costs (or higher fitness scores) than correct matches. To filter out these mismatches, the consistency check is often applied. As shown in Figure \ref{fig:overview}, our approach alternates between the depth/normal estimation stage and the consistency check stage until the whole process converges. That is, once normal and depth map calculation is completed for all views, the algorithm will cross-check the obtained depth/normal among these views. The solutions that pass the check will be marked as validated for the next calculation cycle. It is worth noting that validated and unvalidated solutions are continued to be refined in future optimizations. Unlike unvalidated solutions, validated ones are used for: 1) enforcing smoothness constraint during intra-image solution propagation; 2) propagating solutions among different views; 3) providing visibility information during future source view selection.

Ideally, if a solution $\langle d_i(x), \vec{n}_i(x) \rangle$ for pixel $x$ in image $I_i$ is correct, it should be consistent with all corresponding solutions in the source views used for pixel $x$. That is, after applying homography transform $H^{ij}$ between view $I_i$ and one of the source view $I_j$ for pixel $x$, we should have  $H^{ij}(d_i(x)) = d_j(H^{ij}(x))$ and $H^{ij}(\vec{n}_i(x)) = \vec{n}_j(H^{ij}(x))$. In practice, we consider solution $\langle d_i(x), \vec{n}_i(x) \rangle$ is consistent with source view $I_j$ if the following two criteria hold:
\begin{eqnarray}
    |H^{ij}(d_i(x))-d_j(H^{ij}(x))| &<& {\mathit{T}}_{depth} , \\
    \arccos(H^{ij}(\vec{n}_i(x)) \cdot \vec{n}_j(H^{ij}(x))) &<& {\mathit{T}}_{normal} \nonumber 
\end{eqnarray}
where ${\mathit{T}}_{depth}$ and ${\mathit{T}}_{normal}$ are two preset thresholds (it is set to 0.01 mm and 30 degrees in practice). In addition, instead of requiring all source views used for pixel $x$ to be consistent with $\langle d_i(x), \vec{n}_i(x) \rangle$, we relax the constraint by allowing a small percentage of these views do not satisfy the above conditions. That is, $\langle d_i(x), \vec{n}_i(x) \rangle$ is labeled as verified if it is consistent with 70\% or more of the source views $\{I_{src}\}$ selected for pixel $x$. %\zt{Note that $\{I_{src}\}$ is already been processed for every pixel.} 

\subsection{Propagation Between Views}
\begin{figure}[t]

    \includegraphics[width=\linewidth]{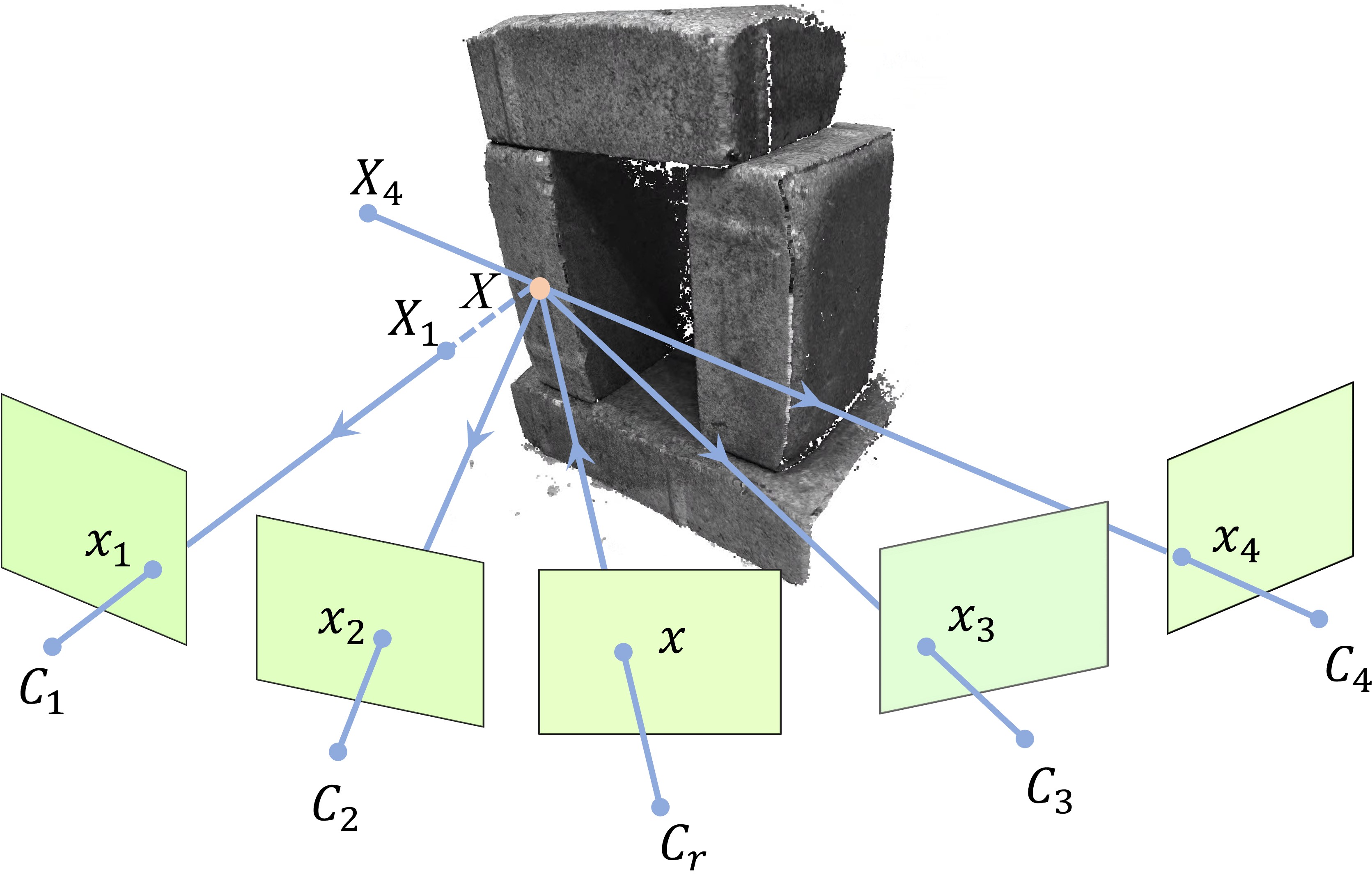}
    \caption{Illustration of the Propagation Between Views. The solution in reference view $C_r$ is consistent with two of its neighboring views: $C_2,C_3$. While the rest of the two are not consistent with $C_r$. Then the solution will be propagated to $x_1, x_4$ to check whether it is a better solution to replace existing ones. }
    \label{fig:prop}

  \hfill
\end{figure}

Initially, solution propagation only appears in neighboring colonies within the same image (referred as intra-image propagation) via onlooker bees. It is one of the vital concepts in PatchMatch-based multi-view stereo. In this paper, we propose inter-image propagation as well, which enables the solution to propagate between pixels in different images related by consistency check. Figure \ref{fig:prop} shows an example of propagation between views. The key concept is to propagate each solution that is consistent with at least one other source view to views that do not have validated solutions at the corresponding locations. This helps to speed up the convergence and prevent potential local optimum.

\subsection{Smoothness Constraint}
Not being able to handle textureless areas properly is a significant limitation for the PatchMatch-based methods \cite{romanoni2019tapa}. In binocular stereo, this problem is often addressed by introducing an additional smoothness term, which converts per-pixel optimization into global optimization. Solving global optimization under the MVS setting can be highly computationally expensive. Hence, we applied a simple yet effective approach, which adds fitness reward to solutions propagated by onlooker bees. The key observation is that the fitness values for the correct match and mismatches are similar in textureless areas. Adding a small reward to the validated solutions propagated by onlooker bees effectively encourages the same fitting plane being selected at the current solution. The smoothness is therefore enforced, and flat textureless surfaces can be properly modeled. For areas with distinct textures, the small reward will not affect the search for optimal solutions.

\section{Fusion}\label{sec:fusion}

Follows \cite{galliani2015massively, schonberger2016pixelwise}, after obtaining all the depth and normal maps, we fuse them into a single point cloud. More specifically, for $N$ images in the scene, we consequently select each image as the reference image and convert its depth map to 3D points in the world coordinate, then project them to the rest $N-1$ views. If the relative depth difference is less than 0.01 mm, and the angle between normals is smaller than 30 degrees, then it is counted as a consistent view. If there exist more than three consistent views, then the point will be accepted in the result. Finally, the points that are related by the depth and normal estimates are averaged into a 3D point in the result point cloud.
\begin{table}
  \centering
  \begin{tabular}{p{0.3\linewidth}p{0.25\linewidth}p{0.25\linewidth}p{0.2\linewidth}}
    \toprule
    Method & Acc.(mm)  &   Comp.(mm)  &  Overall \\
    \midrule
    Furukawa \cite{furukawa2009accurate} & 0.605 & 0.842 & 0.724 \\
    Tola \cite{tola2009daisy} & 0.307 & 1.097 & 0.702 \\
    COLMAP \cite{schonberger2016pixelwise} & 0.400 & 0.532 & 0.664 \\
    Campbell \cite{campbell2008using} & 0.753 & 0.540 & 0.647 \\
    Gipuma \cite{galliani2015massively} & \textbf{0.273} & 0.687 & 0.480 \\
    Ours & 0.385 & \textbf{0.388} & \textbf{0.386} \\
    \bottomrule
  \end{tabular}
  \caption{Quantitative results for non-learning-based approaches on \textbf{full} DTU dataset. Lower is better. Our method ranks first in terms of Completeness and Overall metrics. }
  \label{tab:non-learning}
\end{table}

\begin{table}
  \centering
  \begin{tabular}{p{0.4\linewidth}p{0.2\linewidth}p{0.25\linewidth}p{0.15\linewidth}}
    \toprule
    Method & Acc.(mm)  &   Comp.(mm)  &  Overall \\
    \midrule
    \textbf{Non-Learning-based} \\
    Furukawa \cite{furukawa2009accurate} & 0.613 & 0.941 & 0.777 \\
    Tola \cite{tola2009daisy} & 0.342 & 1.190 & 0.766 \\
    Campbell \cite{campbell2008using} & 0.835 & 0.554 & 0.695 \\
    Gipuma \cite{galliani2015massively} & \textbf{0.283} & 0.873 & 0.578 \\
    COLMAP \cite{schonberger2016pixelwise} & 0.411 & 0.657 & 0.534 \\
    Ours & 0.405 & \textbf{0.381} & \textbf{0.393} \\
    \hline
    \textbf{Learning-based}\\
    SurfaceNet \cite{ji2017surfacenet} & 0.450 & 1.040 & 0.745 \\
    MVSNet \cite{yao2018mvsnet} & 0.396 & 0.527 & 0.462 \\
    P-MVSNet \cite{luo2019p} & 0.406 & 0.434 & 0.420 \\
    R-MVSNet \cite{yao2019recurrent} & 0.383 & 0.452 & 0.417 \\
    CasMVSNet \cite{gu2020cascade} & \textbf{0.325} & 0.385 & 0.355 \\
    PatchMatchNet \cite{wang2021patchmatchnet} & 0.427 & \textbf{0.277} & 0.352 \\
    UniMVSNet \cite{peng2022rethinking} & 0.352 & 0.278 & \textbf{0.315} \\
    
    \bottomrule
  \end{tabular}
  \caption{Quantitative results on DTU \textbf{evaluation set}. Both learning-based and non-learning-based approaches are listed for impartial comparison. }
  \label{tab:learning}
\end{table}

% \begin{table*}

% \begin{tabular*}{\textwidth}{@{\extracolsep{\fill}}l c c c c c c c c c c c }
%     \toprule
%     \multirow{2}{*}{Method} & \multicolumn{3}{c}{Artificial Multi-Bee Colony} & \multirow{2}{*}{SC} & \multirow{2}{*}{PVS} &  \multirow{2}{*}{PVB} & \multirow{2}{*}{Acc. } & \multirow{2}{*}{Comp. } & \multirow{2}{*}{Overall}\\
%     \cline{2-4}
%     & Employed. & Onlooker.& Scout.&\\
%     % Method & & Accuracy &   Completeness & Overall \\
%     \midrule

%     Baseline & \checkmark & \checkmark & \checkmark &&& & 0.367 & 0.451 & 0.409 \\
%     Baseline + PVS + PBV & \checkmark & \checkmark & \checkmark &\checkmark&\checkmark& & 0.367 & 0.451 & 0.409 \\
%     Baseline + SC + PVS + PBV & \checkmark & \checkmark & \checkmark & \checkmark & \checkmark & \checkmark & 0.405 & 0.381 & 0.393 \\
%     \bottomrule
% \end{tabular*}
%   \caption{Ablation Study on DTU dataset. Lower is better.}
%   \label{tab:ablation}

% \end{table*}

\section{Experiments}

We evaluate our method on the DTU Robot Image dataset \cite{aanaes2016large}. In this section, we present the dataset details, evaluation results, ablation study for key components, and implementation details. Noted that $\mathit{FoodNum}$ is set to 10 in our experiments.

\subsection{DTU Robot Image Dataset}

As our main testing dataset, the DTU dataset contains 124 different scenes captured by a structured light scanner mounted on an industrial robot arm. Each scene has been taken from 49 or 64 positions with seven different lighting conditions. In this paper, we select the most diffuse set. The image resolution is $1600 \times 1200$, and the camera calibration parameters are provided.\\

% \noindent\textbf{Middlebury MVS Dataset}

% The Middlebury MVS dataset contains two different objects in three settings: full, ring, and sparse ring. The full setting has 312 to 363 views. The ring setting has 47 to 48 views. And the sparse ring setting has 16 views. We only focus on ring and sparse ring settings, since full setting is a bit redundant in the number of views. The image resolution is 640 x 480. And the camera calibration parameters are provided.

\begin{figure*}
        % \begin{subfigure}[]{0.3333\textwidth}
        %         \includegraphics[width=\linewidth]{cvpr2023-author_kit-v1_1-1/latex/imgs/95-100-cycle0.jpg}
        %         \caption{Ours without Smoothness Constraint}
        %         \label{fig:gull}
        % \end{subfigure}%
        % \begin{subfigure}[]{0.3333\textwidth}
        %         \includegraphics[width=\linewidth]{cvpr2023-author_kit-v1_1-1/latex/imgs/95-100-cycle15.jpg}
        %         \caption{Ours}
        %         \label{fig:gull2}
        % \end{subfigure}%
        % \begin{subfigure}[]{0.3333\textwidth}
        %         \includegraphics[width=\linewidth]{cvpr2023-author_kit-v1_1-1/latex/imgs/95-100-gt.jpg}
        %         \caption{Ground Truth}
        %         \label{fig:gull2}
        % \end{subfigure}%

                \includegraphics[width=\linewidth]{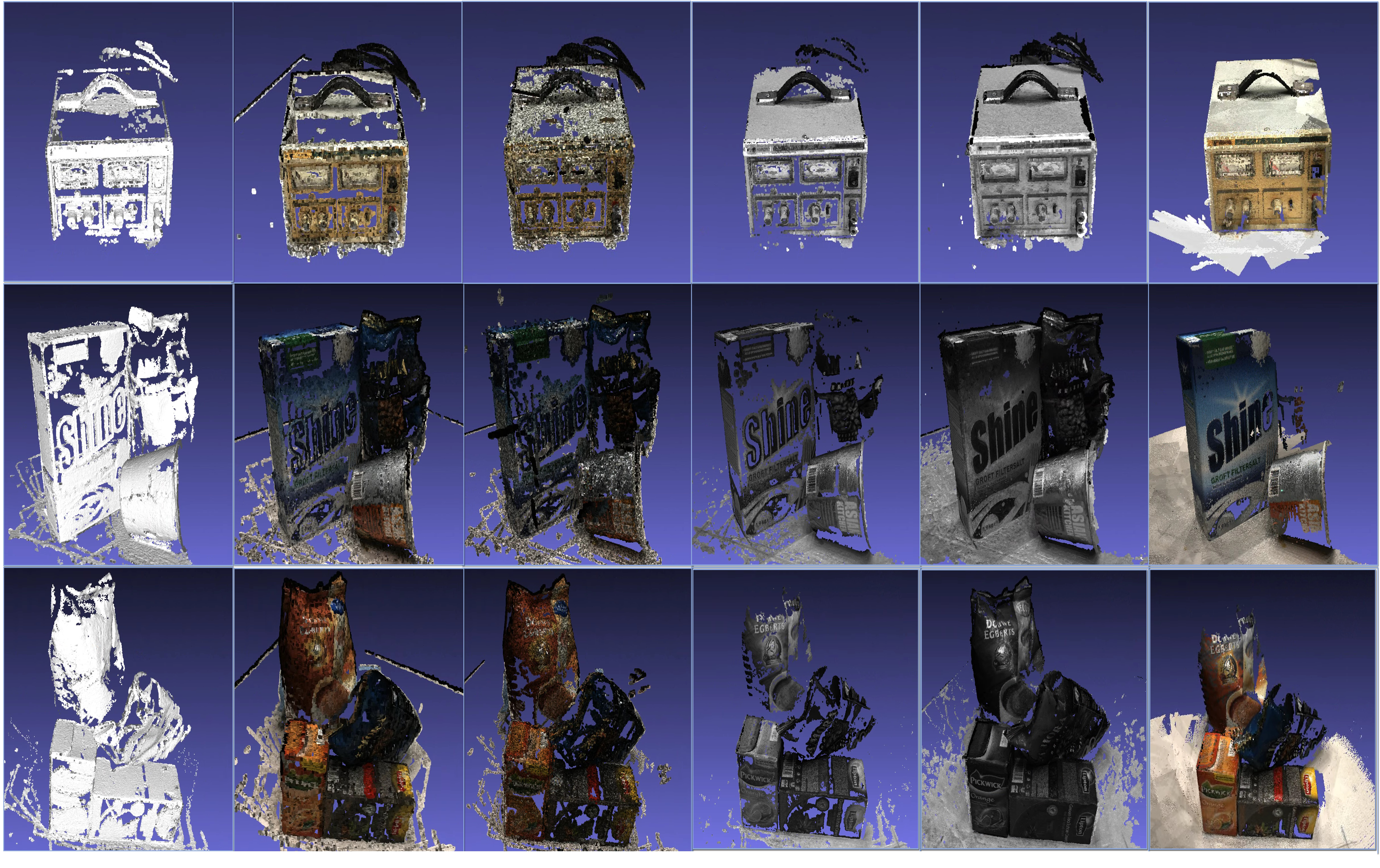}
        \caption{Visualization results of the proposed method in scene 11, 95 and 100 in DTU dataset. From left to right: Tola \cite{tola2009daisy}, Furukawa \cite{furukawa2009accurate}, Campbell \cite{campbell2008using}, point cloud without Smoothness Constraint, point cloud with Smoothness Constraint, ground truth.}\label{fig:abalation-smooth}
\end{figure*}

\begin{figure*}
    \centering
    \includegraphics[width=\textwidth]{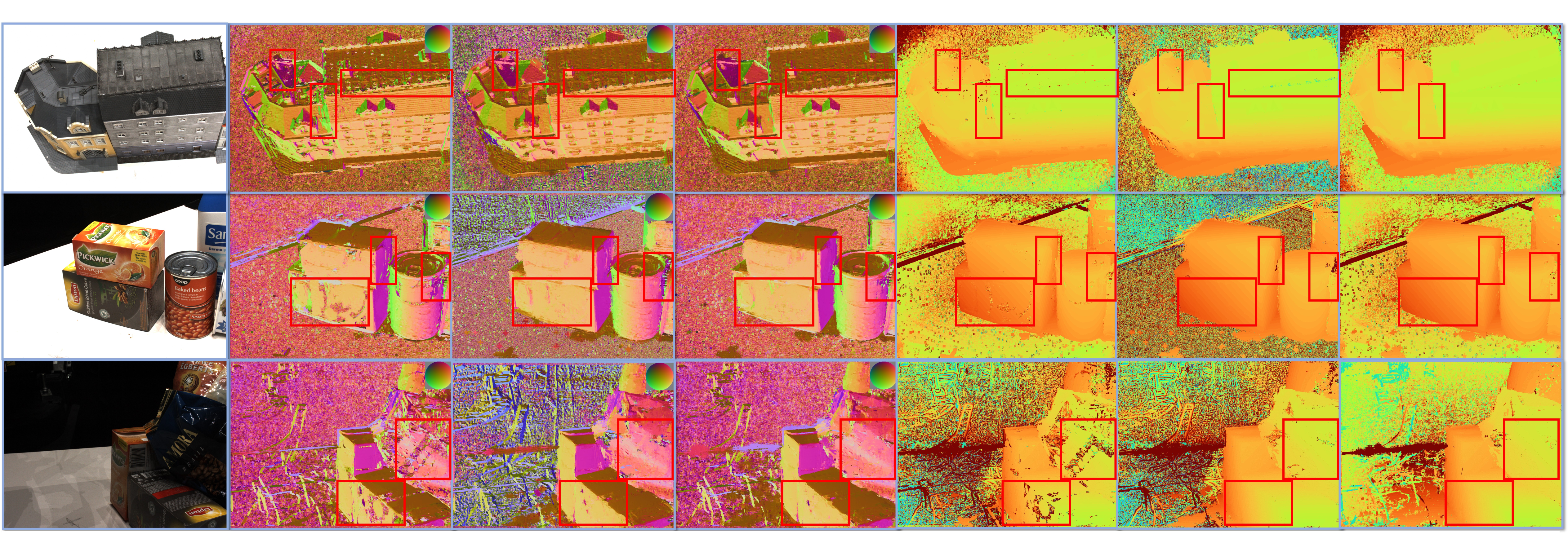}
    \caption{Illustration on the effect of the Smoothness Constraint (SC) and Pixelwise View Selection (PVS). From top to bottom: scene 19, 59, 100 from the DTU dataset. From left to right: input image, normal map without SC \& PVS, normal map with SC, normal map with SC \& PVS, depth map without SC \& PVS, depth map with SC, depth map with SC \& PVS. Challenging areas are marked in red boxes.}
    \label{fig:pvs-sc}
\end{figure*}

\subsection{Point Cloud Evaluation}

% \noindent\textbf{DTU Dataset}

% For point cloud evaluation, we strictly follow the authors' instructions. 
Accuracy is measured as the distance from the reconstruction point cloud to the structured light reference. Completeness is measured from the structured light reference to the reconstruction point cloud. The overall score is computed by averaging the accuracy and completeness. Note that the reconstruction point cloud is downsampled to ensure unbiased evaluation since strongly textured regions generally have dense 3D points. The evaluation program is provided by the authors.

For this dataset, we present two versions of quantitative results: one for comparison of the non-learning-based methods and the other for comparison of both the learning-based and non-learning-based methods. For non-learning-based method evaluation, we follow the protocol specified by the authors of the dataset by testing 80 different scenes. For Learning-based method evaluation, we follow \cite{yao2018mvsnet} by using the validation set containing 22 different scenes for impartial comparison. In Table \ref{tab:non-learning}, we present the quantitative results for non-learning-based approaches on the full DTU dataset. Our approach performs the best in both completeness and overall metrics. In Table \ref{tab:learning}, we present the quantitative results for both learning-based and non-learning-based methods on the DTU evaluation set. Our method shows competitive performance to the learning-based methods. \\

% \textbf{Middlebury MVS Dataset}

\subsection{Ablation Study}

We here present the ablation study of the proposed method. The baseline method is generated using AMBC with all three types of bees: employed bees, onlooker bees, and scout bees. Our proposed baseline method with smoothness constraint, pixelwise view selection, and propagation between views shows the best performance. Note that propagation between views only speeds up the convergence because it does not modify the solution space. Therefore it does not have a significant impact on result accuracy.

Figure \ref{fig:pvs} demonstrates the effect of the pixelwise view selection in occluded regions. The pixel-wise view selection could handle the occluded regions and generate correct normal and depth estimation. Figure \ref{fig:abalation-smooth} shows a comparison of the point clouds on scene 11, 95, and 100 in the DTU dataset. The proposed method with smoothness constraint clearly outperforms in low-textured areas. In addition, only a few noisy/incorrect points are introduced along the edges of the surfaces. Figure \ref{fig:pvs-sc} illustrates the effect of the combination of smoothness constraint and pixelwise view selection. The best results are obtained by applying both of them. The numerical results of the ablation study are provided in Supplementary Materials.

\section{Conclusion}
In this paper, we present a visibility-aware pixelwise view selection method for PatchMatch-based multi-view stereo. View selection is progressively improved for individual pixels as more knowledge of scene geometry is obtained. Selected views are used for both matching cost evaluation and consistency check. When applying Artificial Multi-Bee Colony (AMBC) to search optimal solutions for different pixels in parallel, between-colony onlooker bees are used for intra-image and inter-image solution propagation. To tackle the lack of photometric cues at low-textured regions, fitness rewards are introduced on those solutions verified by consistency check. Experiments on the DTU dataset demonstrate that our method achieves state-of-the-art performance among the non-learning-based methods. Ablation study shows that our two main components, visibility-aware pixelwise view selection, and smoothness rewards, can notably improve the handling of occluded and low-textured areas. The source code will be released after the acceptance of the paper.

%%%%%%%%% REFERENCES
{\small
\bibliographystyle{ieee_fullname}
\bibliography{egbib}
}

\end{document}